\documentclass[journal]{IEEEtran}
\ifCLASSINFOpdf
\else
\fi
\usepackage{graphicx}
\usepackage{amsmath}
\usepackage{longtable}
\usepackage{color}

\usepackage{algorithm}
\usepackage[noend]{algpseudocode}
\usepackage{booktabs}
\begin{document}
%
\title{A General Data Renewal Model for Prediction Algorithms in Industrial Data Analytics}
%
%
%

\author{Hongzhi~Wang,~\IEEEmembership{Member,~IEEE,}
        Yijie~Yang,
        and~Yang~Song
\thanks{Hongzhi Wang, Yijie Yang and Yang Song are with the Department
of Computer Science and Technology, Harbin Institute of Technology, Harbin,
Heilongjiang, 150000 China e-mail: wangzh@hit.edu.cn.}}

\maketitle

\begin{abstract}
In industrial data analytics, one of the fundamental problems is to utilize the temporal correlation of the industrial data to make timely predictions in the production process, such as fault prediction and yield prediction. However, the traditional prediction models are fixed while the conditions of the machines change over time, thus making the errors of predictions increase with the lapse of time. In this paper, we propose a general data renewal model to deal with it. Combined with the similarity function and the loss function, it estimates the time of updating the existing prediction model, then updates it according to the evaluation function iteratively and adaptively. We have applied the data renewal model to two prediction algorithms. The experiments demonstrate that the data renewal model can effectively identify the changes of data, update and optimize the prediction model so as to improve the accuracy of prediction.
\end{abstract}


%
\IEEEpeerreviewmaketitle

\section{Introduction}
In the industrial processes, though strict regulations and stable operating conditions are required, even the most sophisticated machines cannot avoid the runtime exception~\cite{mckee2014review}. Due to a large number of human interactions in the industrial manufacturing process, human errors may cause abnormalities in the overall process. According to statistics, the maintenance cost of various industrial enterprises accounts for about 15\%-70\% of the total production cost \cite{bevilacqua2000analytic}. Therefore, how to conduct malfunction analysis, yield dynamic prediction and make timely fault detection for industrial processes to ensure the effectiveness and efficiency of the production process have received much attention in both academia and industry.

Because of the numerous sensors with high sampling frequency in industrial processes, the devices will accumulate large amounts of data in a short time interval. As time goes on, some parameters related to production prediction and fault detection are imperative to change synchronously due to the equipment aging, abrasion and so on.  However, the currently known prediction algorithms in industry, mainly including artificial intelligence and data-driven Statistical methods\cite{pecht2009prognostics} \cite{sikorska2011prognostic}\nocite{huang2017remaining}\nocite{wang2016residual}\nocite{si2012remaining}\nocite{zhuang2016fault}\nocite{wan2017manufacturing}\nocite{hasani2017automated}\nocite{zhang2017opportunistic}\nocite{batzel2009prognostic}\nocite{kharoufeh2013reliability}
, are all constrained by time, up to a point. In other words, these prediction models can only accurately reflect the state of the industrial equipment in a certain period, whereas the inaccuracy increases over time.

Additionally, in the problems of malfunction diagnoses and predictions based on transfer learning\cite{Long2015Learning}\cite{Ran2017Transfer}, though there are various kinds of faults in industrial processes, the similarities of them can be utilized to conduct the transfer learning of the malfunctions, so as to predict the faults efficieintly and effectively. We will give an outline of a transfer-learning-based fault prediction algorithm in Section 3. However, in our experiments, we find that the transferability of two different types of equipment in the same technological process considerably decreased if the data used in the process are at different periods. It shows that the crucial precondition for the application of transfer learning to industrial time-series data is that the data of the two different types of industrial equipment in the same production process should also be in the same period. Therefore, the prediction model is required to recognize the changes of industrial data stream and update the parameters automatically with the lapse of time.

This paper focuses on the automatic update and replacement of the model based on industrial time-series data, which are featured with periodicity and complex correlation. To address these problems, this paper proposes a general data renewal model  based on lifelong machine learning\cite{chen2016lifelong}\cite{liu2017lifelong}\cite{silver2013lifelong}.\nocite{balcan2015efficient}\nocite{verstaevel2017lifelong}\nocite{silver1996parallel}\nocite{singh1992transfer}\nocite{silver2002task} It can be applied to some prediction algorithms to improve their accuracy. 
The main idea of the model is to assess the freshness of the industrial time-series data according to the existing prediction model and the new data stream. Then it can decide whether to invalid the old model and retrain a new one according to the similarity function and the loss function.

In our previous work, we attempted to establish a time-series forecasting model system which could solve the problems of both discrete and continuous variable prediction. We have proposed a time-series yield prediction algorithm and a transfer-learning-based fault prediction algorithm, which will be roughly described in Section 3. However, their practical effects were limited due to the reasons mentioned above.  Therefore, in our experiments, we will apply the data renewal model to them and testify its effectiveness by making a comparison between the learning models with and without a data renewal model.

This paper makes following contributions.
\begin{itemize}
	\item We propose a general data renewal model combined with the similarity function and the loss function. It can be applied to some industrial prediction algorithms to find the regulations of renewing the prediction model based on industrial time-series, so as to update the prediction model opportunely and iteratively.
	
	\item Through self-learning and automatic updating, the prediction algorithms applied with the data renewal model can be improved over time, thus reducing human interventions and perfecting the algorithm performance in industrial processes.
	
	\item We evaluate the proposed model on three datasets and two prediction algorithms in industry. The results demonstrate that the model can be updated effectively according to the industrial data stream, and the accuracy of the predictions can be increased by at least 33\%, which is a significant improvement.
\end{itemize}

The rest of the paper is organized as follows. In Section 2, firstly, we define the problem and our target, then describe the model and approach in detail. And we apply our data renewal model to two prediction algorithms. The tuning process,  brief introduction of the two prediction algorithms and the experimental results are presented in Section 3.  Section 4 makes a summary of the paper; meanwhile, it explains the future work.


\section{Prediction Algorithms combined with Data Renewal Model}

\subsection{Task Definition and Overview}
In a prediction algorithm, we build a renewal model based on the data, then continuously update the prediction model according to the data similarity and the loss function. In practical problems, the inputs are often time series. Given a sequence of data points measured in a fixed time interval, $X_{t} = [x_{t1} \quad x_{t2} \quad \cdots x_{tm} ]\in R^{t\times m} $,  and the existing model $M$. The output is the updated prediction model $M'$.

The problem is formalized as follows. Given the data of a piece of industrial equipment $E$ acquired from time $0$ to time $t$, for model $M$, there exists a function $f(\cdot )$,  such that

\begin{equation}
	M(X_t,f(\cdot))= Y_t
\end{equation}

In the meantime, $E$ keeps on running. Given the data of $E$ from time $t$ to time $t+n(n>0)$, $X_{t+n} = [x_1 \quad x_2 \cdots x_m ]\in R^{n\times m} $, for model $ M'$, there exists a function $f'(\cdot )$,  such that
\begin{equation}
	M'(X_{t+n},f'(\cdot))= Y_{t+n}
\end{equation}

If $f(\cdot)=f'(\cdot)$, the model does not need to update. If $f(\cdot)\not=f'(\cdot)$, the model needs further analysis to determine whether it should be updated. If it is, then let $M =M'$.

In practical industrial problems, $X$ are time series. Therefore, we need to choose the model according to their features. Since the model often has a complicated mechanism and the correlation may be nonlinear, $f$ is difficult to find an analytical solution. To achieve better performance, we often use neural network algorithms, such as BP Neural Network\cite{Liu2017A}, Convolutional Neural Network\cite{Shin2016Deep} and Recurrent Neural Network (LSTM)\cite{Ma2015Long}

In most cases, the loss function is enough to measure and determine whether a model needs to update. However, since the industrial data are often discrete, continuously collected and transferred by the sensors,  the traditional mechanism model is ineffective. As a result, the model should be based on data instead of mechanism. For accurate estimation, the similarity function and the loss function are combined to estimate the time of updating the existing model automatically, then the model is retrained on the calculated time iteratively.

\subsection{Data Renewal Model}


\begin{figure}[h]
	\centering  
	\includegraphics[width=1.0\linewidth ]{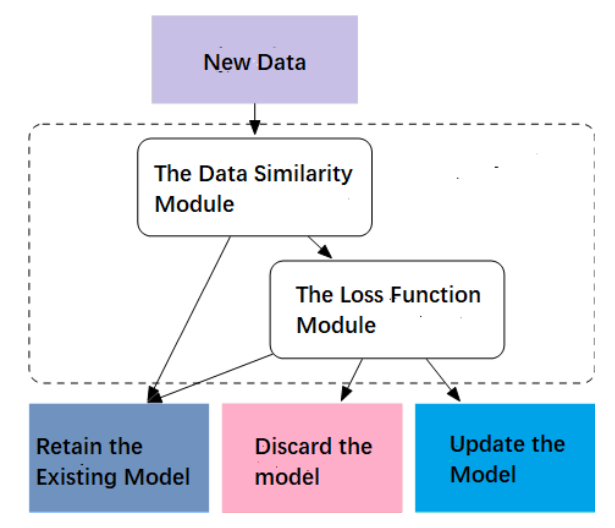}  
	\caption{The Components of the Data Renewal Model }  
	\label{fig:3}   
\end{figure}

The primary problem of the updating algorithm is to build a data renewal model to predict when the model ought to be updated. As shown in Figure~\ref{fig:3}, the model achieves this by two considerations. One is the similar of the original data and the new data. If they are similar enough, the model does not have to be updated or retrained. It is measure by the similarity computation, which will be described in Section~\ref{sec:sim}.
The other is the applicability of the existing model. If the model loses efficacy in new data stream, it should be updated or even retrained. It is measured by the loss function, which will be described in Section~\ref{sec:loss}.

\subsubsection{Similarity Computation}
\label{sec:sim}

The data similarity computation begins with analyzing the changed extent of the unprocessed data in the previous moment with the data in the next moment. Chiefly, the data are classified into two types: the binary attribute data and the numeric data. They will be discussed in this section, respectively.

For the binary attribute data, the similarity is measured by the counting method. That is, for time series $X_{t} = [x_{t1} \quad x_{t2} \cdots x_{tm} ]\in R^{t\times m} $ and $X_{t+n} = [x_1 \quad x_2 \cdots x_m ]\in R^{n\times m} $, assuming that $X_t^i$ and $X_{t+n}^i$, the $i$-th dimension of $X_t$ and $X_{t+n}$, are both binary attributes, the number of data pairs of them both belong to the first state is $S1$, while the number of data pairs are both the second state is $S2$, the total amount of data is $n$, then their similarity is shown as follows.

\begin{equation}
	sim(X_t^i,X_{t+n}^i)=\frac{S_1+S_2}{n}
\end{equation}
For a specific example, $X_t^i=[1~0~0~0~1~0~1~1]$, $X_{t+n}^i=[0~0~0~1~1~1~1~1]$, the 2nd and 3rd numbers of $X_t^i$ and $X_{t+n}^i$ are both 0, so $S1=2$, the 5th, 7th and 8th numbers of $X_t^i$ and $X_{t+n}^i$ are both 1, so $S2=3$, $sim(X_t^i,X_{t+n}^i)=\frac{2+3}{8}=0.625$.

In the case of the numerical data, generally, there are two methods to measure the similarity degree: similarity coefficient and similarity measurement. Pearson Correlation Coefficient, as shown in equation (4), can be used to avoid the error of similarity measurement resulting from the severe dispersion of industrial data. It has a value between $+1$ and $-1$, where $+1$ is total positive linear correlation, 0 is no linear correlation, and $-1$ is total negative linear correlation.
\begin{equation}
	\begin{aligned}
		\rho_{X,Y}=\frac{cov(X,Y)}{\sigma_{X}\sigma_{Y}}
		=\frac{E[(X-\mu x)(Y-\mu y)]}{\sigma_{X}\sigma_{Y}}
	\end{aligned}
\end{equation}

For the data in the same set but different periods, they also have a certain degree of similarity. In order to better measure the similarity between them, we take the absolute value and ignore the positive and negative correlation.
In conclusion, for the purpose of achieving better performance, we adopt the similarity coefficient method here and modify the Pearson Correlation. The new coefficient formula is shown as follow.

\begin{equation}
	\scriptsize
	\begin{aligned}
		sim(X_t^i,X_{t+n}^i) &=\rho_{X^i_t,X^i_{t+n}}=|\frac{cov(X_t^i,X^i_{t+n})}{\sigma_{X^i_t}\sigma_{X^i_{t+n}}}| \\
		&=|\frac{E[(X_t^i-\mu x_t^i)(X^i_{t+n}-\mu x^i_{t+n})}{\sigma_{X^i_t}\sigma_{X^i_{t+n}}}|
	\end{aligned}
\end{equation}

With respect to different data vectors in the same data set, the similarity is gained by calculating the mean value of them and adjusted overall using the parameter $\delta$, which is shown in (5). In most cases, $\delta_i=1$, if the data set does not contain attribute $i$, then $\delta_i=0$.

\begin{equation}
	\scriptsize
	\begin{aligned}
		sim(X_t,X_{t+n}) &=1-\frac{\sum^m_{i=1}\delta_i(1-sim(X_t^i,X^i_{t+n})}{\sum^m_{i=1}\delta_i }\\
		&=\frac{\sum^m_{i=1}\delta_i\cdot sim(X_t^i,X^i_{t+n})}{\sum^m_{i=1}\delta_i }
	\end{aligned}
\end{equation}

According to (3), (4) and (5), it can be concluded that the value of the similarity is in the range $(0, 1)$. The higher the value is,  the higher the degree of similarity should be between the two data sets. When the value of the similarity function is lower than a threshold $z$, the model needs to be updated. 

\subsubsection{Loss Function}
\label{sec:loss}
The loss function aims to determine whether to update or abandon the old model and train a new one. In fact, the updating is to adjust the parameters in the existing model. Therefore, it is necessary to estimate the effect of the model, and generally it is measured by loss function.


\begin{equation}
	\theta \ast =argmin \frac{1}{N}\sum_{i=1}^{N}L(y_i,f(x_i;\theta _i))+\lambda \Phi(\theta)
\end{equation}
where $L$ is the loss function, $ \Phi(\theta)$ is the regularization term or the penalty term. As regards to specific problems, we use a more specific loss function for analysis. One of the typical problems is the updating of the industrial big data model, which mainly involves two aspects, the model for continuous data such as yeild prediction, and the model for categorical attribute data such as fault prediction. According to the features of different models, we should define the corresponding loss function for estimation.

For continuous data prediction, we use RMSE (Root Mean Square Error) to evaluate the loss, since it shows greater estimation results than quadratic loss function in our experiments.

\begin{equation}
	RMSE(X,f(\cdot)) = \sqrt{\frac{1}{m}\sum^m_{i=1}(f(x_i)-y_i)^2}
\end{equation}

For classification prediction problems, 0-1 loss function, Log loss function, Hinge loss, and perceptual loss function\cite{rosasco2004loss}
\cite{shen2005loss}
\cite{masnadi2009design}
are generally used to estimate the quality of the model. Here, we use the perceptual loss function:
\begin{equation}
	\mathop{\arg\min}_{w,b} \ \ [-\sum^n_{i=1}y_i(w^Tx_i+b)]
\end{equation}

In the loss function,  two questions should be taken into account, whether to update the existing model and whether to abolish the old one. The loss of the existing model is $Lm$, then the new data are collected and put into it for training, and the loss of the new model is $Ln$. Then the change rate of the loss is computed as follow.

\begin{equation}
	LC=|\frac{Ln-Lm}{Lm}|
\end{equation}

\begin{figure}[h]
	\centering  
	\includegraphics[width=1.0\linewidth ]{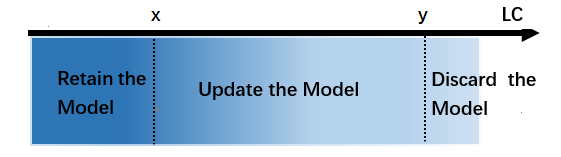}  
	\caption{The State-determining Schematic Diagram of the Loss Function Module  }  
	\label{fig:3-3}   
\end{figure}
In the formula, obviously, $LC$ is greater than 0. The threshold values need to be set to determine when to update and discard the model (retrain a new one). So the they are set as follows: if $LC>y$, the original model will be discarded and a new model will be built through retraining. If $y>LC>x$, the model is updated: new data are added to the model for training, so the model parameters are changed. If $LC<x$, the original model state is retained with no update. The standards are shown in Figure ~\ref{fig:3-3}. The thresholds will be adjusted according to the experiments.

Algorithm 1 describes the procedure of the data renewal model. Line 1 analyzes the data similarity of period $t1$ and period $t2$. Lines 2-11 decide whether to update the model according to the data similarity. Lines 3-9 calculate the value of loss function in the period $t1$ and $t2$ and output the flag bit according to the state-determination rules.

\begin{algorithm}[htb]
	\scriptsize
	\caption{update($data_{T1}$,$data_{T2}$)}
	\label{alg:Framwork}
	\begin{algorithmic} [1]
		\Require
		The data set $data_{T1}$ in period $t1$;
		The data set $data_{T2}$ in period $t2$
		\Ensure
		The updating flag of the model $flag$ \\
		$p \gets sim(data_{T1},data_{T2})$
		\If {$p<z$}
		\State $l \gets loss(data_{T1},data_{T2})$;
		\If{$l>y$}
		\State return 2; \qquad // Discard the model and retrain a new one
		
		\ElsIf{$y>l$ and $l>x$}
		\State return 1; \qquad // Add the new data and update the model
		\Else
		\State return 0; \qquad //  Retain the model
		\EndIf
		\Else
		\State return 0;\qquad  //  Retain the model
		\EndIf

	\end{algorithmic}
\end{algorithm}

\subsection{Algorithm Description}

The data renewal model is based on the data similarity and the loss function. Furthermore, to control the updating frequency, the size of the new data should be controled. That is, the similarity will not be calculated until a certain amount of data has been accumulated. Only when the threshold of the data size is reached, will the loss function of the data be computed, and the corresponding processes of the model be performed.

Algorithm 2 describes the procedure of updating according to Algorithm 1 and the algorithm flow discussed above. Lines 1-4 decide whether enough new data have been accumulated,  and Lines 5-12 decide whether to update the model referring to the data renewal model in Algorithm 1. 

\begin{algorithm}[H]
	\scriptsize
	\caption{lifelong($data_{T1}$,$data_{T2}$)}
	\label{alg:Framwork}
	\begin{algorithmic} [1]
		\Require
		The data set $data_{T1}$ in period $t1$;
		The data set $data_{T2}$ in period $t2$;
		The minimum (threshold) amount of data $L$,the model $M$
		\Ensure
		The model $M_f$ \\
		$n \gets shape(data_{T2})$
		\If {$n<L$}
		\State return $M$;
		\Else
		\State $flag \gets update(data_{T1},data_{T2})$;
		\If{$flag = 1$}
		\State $M_f \gets train(data_{T1},data_{T2})$;
		\EndIf
		\If{$flag = 2$}
		\State $M_f \gets train(data_{T2})$;
		\EndIf
		\If{$flag = 0$}
		\State$M_f \gets M$;
		
		\EndIf
		
		\EndIf
		\State return $M_f$;
	\end{algorithmic}
\end{algorithm}


The complexity analysis of the algorithm is focused on the data renewal model.

\underline{Time Complexity Analysis} Let the data size in the period $t_1$ and $t_2$ be $N$ and $M$, respectively. The cost of similarity computation is $O(MN)$. The loss function add the data in period $t_2$ into the existing model and compares the new loss with the previous one. It only needs one round of calculation, so the time complexity is $O(M)$. Therefore, the overall time complexity is $O(MN+M)=O(MN)$

\underline{Space Complexity Analysis}
Let the needed space of the data in period $t_1$ and $t_2$ be $N$ and $M$, respectively. The similarity matrix costs $O(NM)$. $O(M)$ is required to store the intermediate results calculated by the data in period $t_2$ to calculate its loss function. As a result, the overall spatial complexity is $O(N+2M+MN)$ = $O(MN)$

\section{Experiments and Evaluation}
\subsection{Experimental Environment and Datasets}
The experimental environment and datasets are shown in Table 1. Among them, the industrial boiler dataset and the generator dataset are generated by real-time production system in the Third Power Plant of Harbin. The boiler dataset includes more than 400,000 pieces of data, whose dimension are up to 70, and the main attributes involve time, flow, pressure, and temperature. Moreover, the generator dataset includes more than 80,000 pieces of data, whose dimension are up to 38, and the major attributes include time, speed, power, pressure and temperature.

In this section, the data renewal model is respectively applied to the time-series yield prediction algorithm and the transfer-learning-based fault prediction algorithm to verify its effectiveness in model updating and optimization.

\begin{table}[H]
	\caption{The Experimental Environment and Datasets }
	\scriptsize
	\label{Tab:bookRWCal}
	\centering
	\begin{tabular}{lp{4cm}p{4cm}}
		\toprule[1.3pt]
		
		Machine Configuration & 2.7GHz Intel Core i5 8GB 1867 MHz DDR3r \\
		\midrule
		Experimental Environment &Python 3.6.0;   Tensorflow\\
		\midrule
		Datasets &  Industrial Boiler Dataset;  Industrial Generator Dataset;   Synthetic  Industrial Generator Dataset \\
		\midrule
		Algorithms &   The Time-series Yield Prediction Algorithm;  The Transfer-learning-based Fault Prediction Algorithm  \\
		\bottomrule[1.3pt]
	\end{tabular}
\end{table}
\subsection{The Optimization of the Prediction Algorithm Based on Data Renewal Model}
The data renewal model has two critical parameters, the similarity and the change rate of loss. Since the thresholds may have an impact on accuracy, we are imperative to test their impacts.

We first test the impact of the similarity. The goal of the similarity is to estimate the similar degree of the data at different time. The higher the value is, the more similar the data are. Therefore, the similarity threshold can neither be too low nor too high. Initially, it is set to 0.3, 0.5, and 0.7, while the loss rate threshold is set to 1/0.4. In order to observe the changes with different thresholds more intuitively, we set the original number of tuples to 10,000. The model is assessed whether to be updated according to the flag bit for every 10,000 pieces of data added.

\begin{figure}[htb]
	\centering  
	\includegraphics[width=0.8\linewidth]{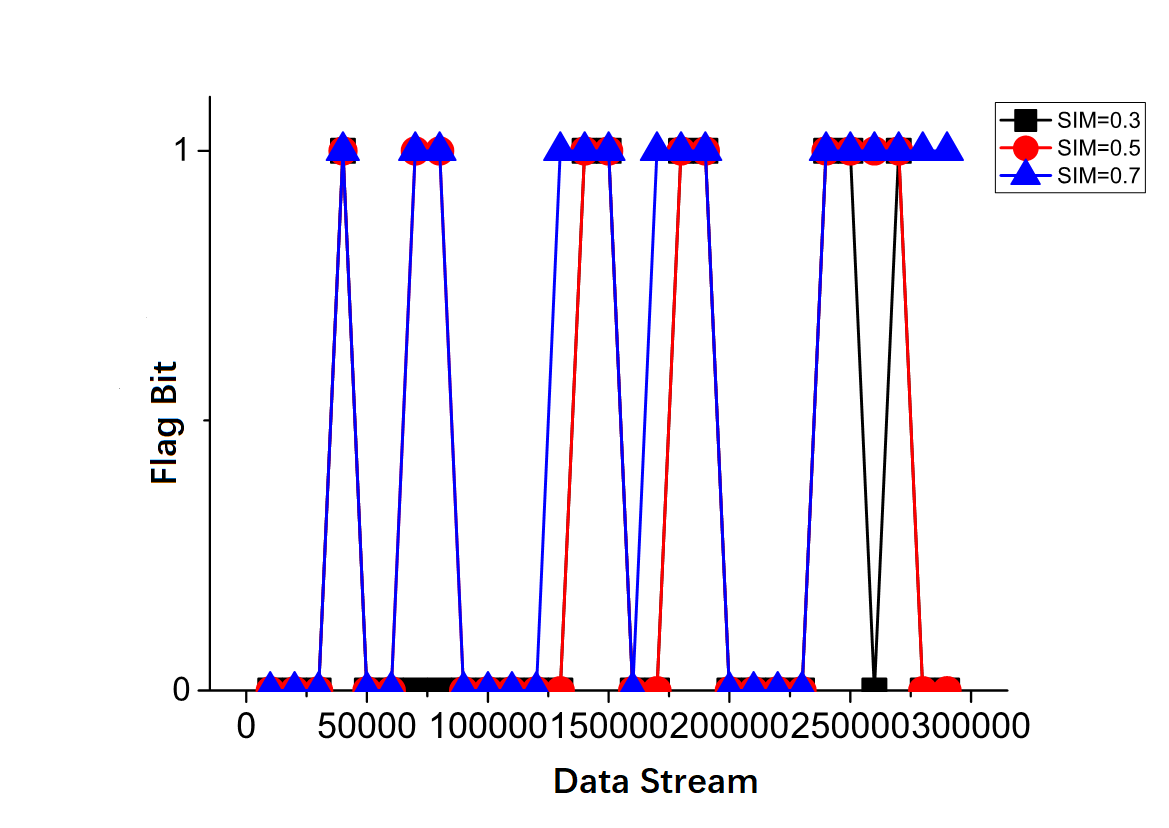}  
	\caption{The Impact of the Similarity for Algorithm 1}  
	\label{fig:6}   
\end{figure}
As shown in Figure~\ref{fig:6}, when the similarity threshold is 0.3, the model updating frequency is low, while the model updates too often when the threshold is 0.7. To ensure that the frequency is kept at an appropriate level, the similarity threshold is set to 0.5 in the next experiments .

\begin{figure}[h]
	\centering  
	\includegraphics[width=0.8\linewidth]{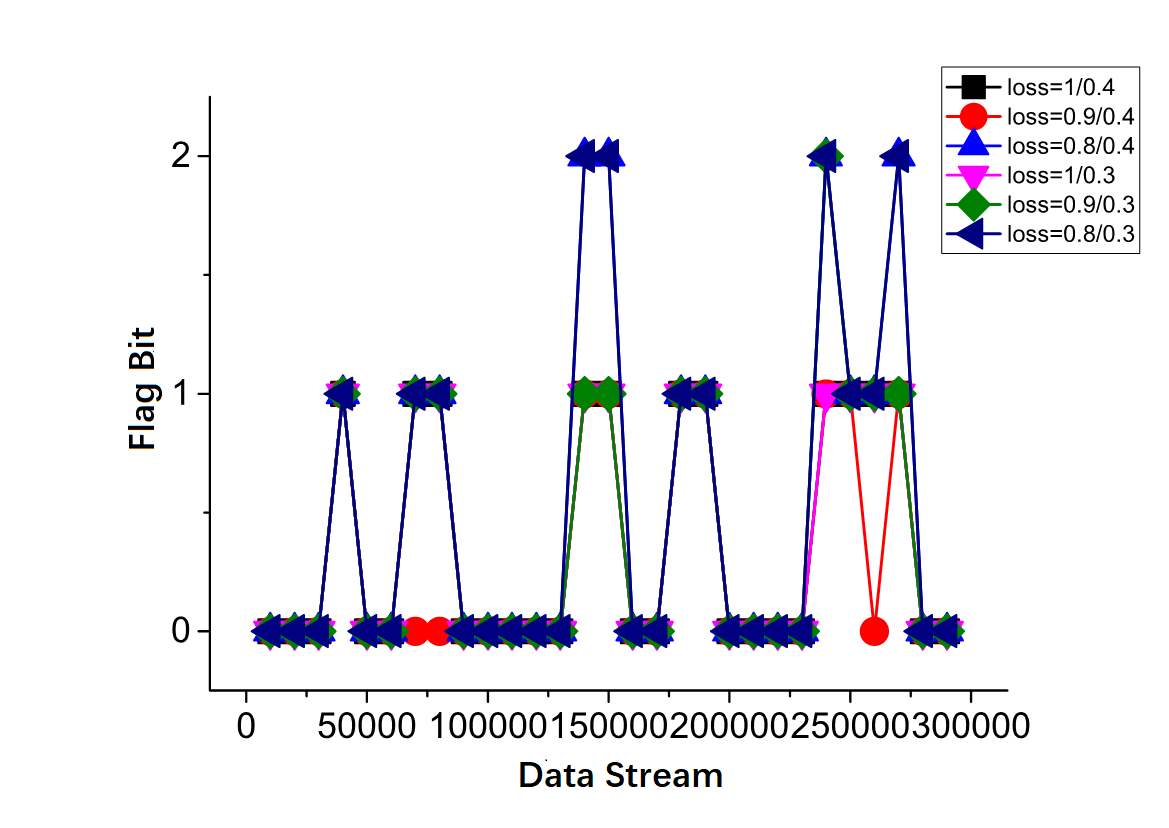}  
	\caption{The Impact of the Change Rate for Algorithm 1}  
	\label{fig:7}   
\end{figure}
Then we test the impact of the change rate of loss. We vary the rate of loss as 1/0.4, 0.9/0.4, 0.8/0.4, 1/0.3, 0.9/0.3, and 0.8/0.3. The results are shown in Figure ~\ref{fig:7}. When the flag is 2, the model is discarded. When the flag is 1, the model is updated, and it is retained if the flag is 0. It is found that the discarding and updating of the model shows a more balanced frequency when the threshold is 0.9/0.3.

\begin{figure}[htb]
	\centering  
	\includegraphics[width=0.8\linewidth ]{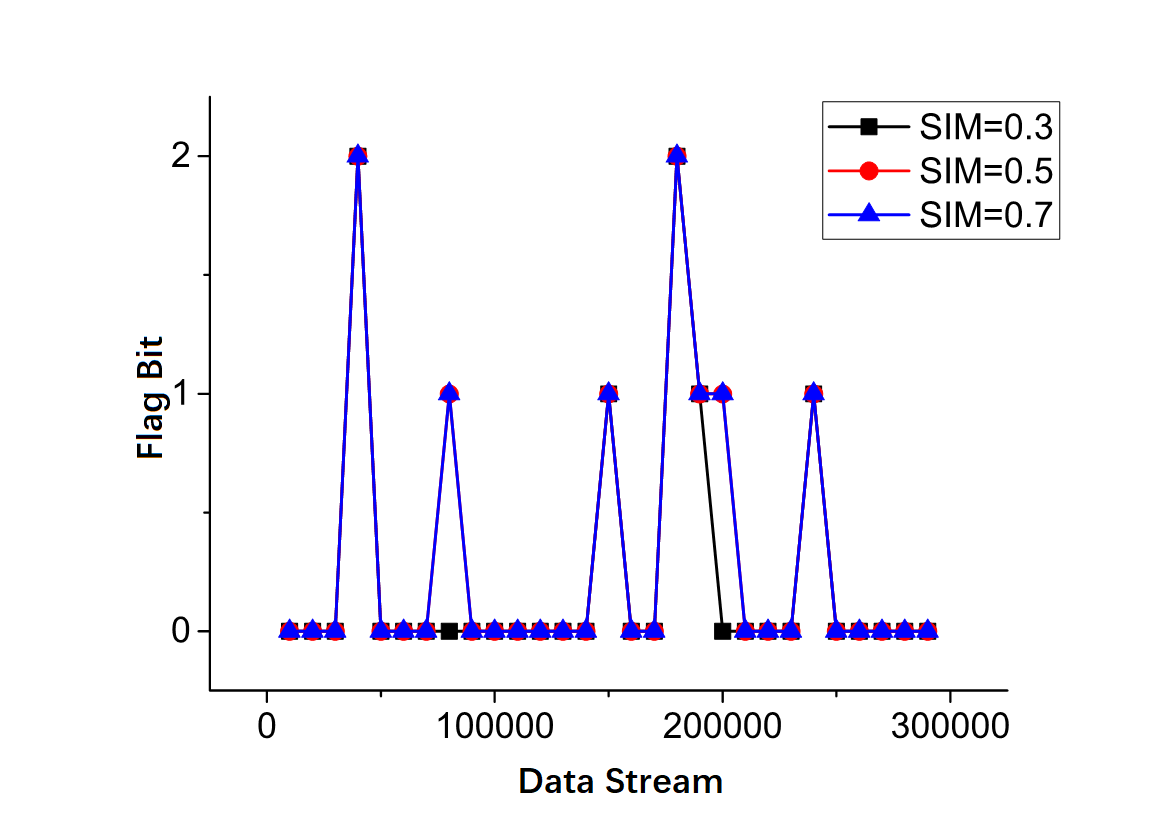}  
	\caption{The Impact of the Similarity for Algorithm 2 }  
	\label{fig:8}   
\end{figure}
\begin{figure}[htb]
	\centering  
	\includegraphics[width=0.8\linewidth ]{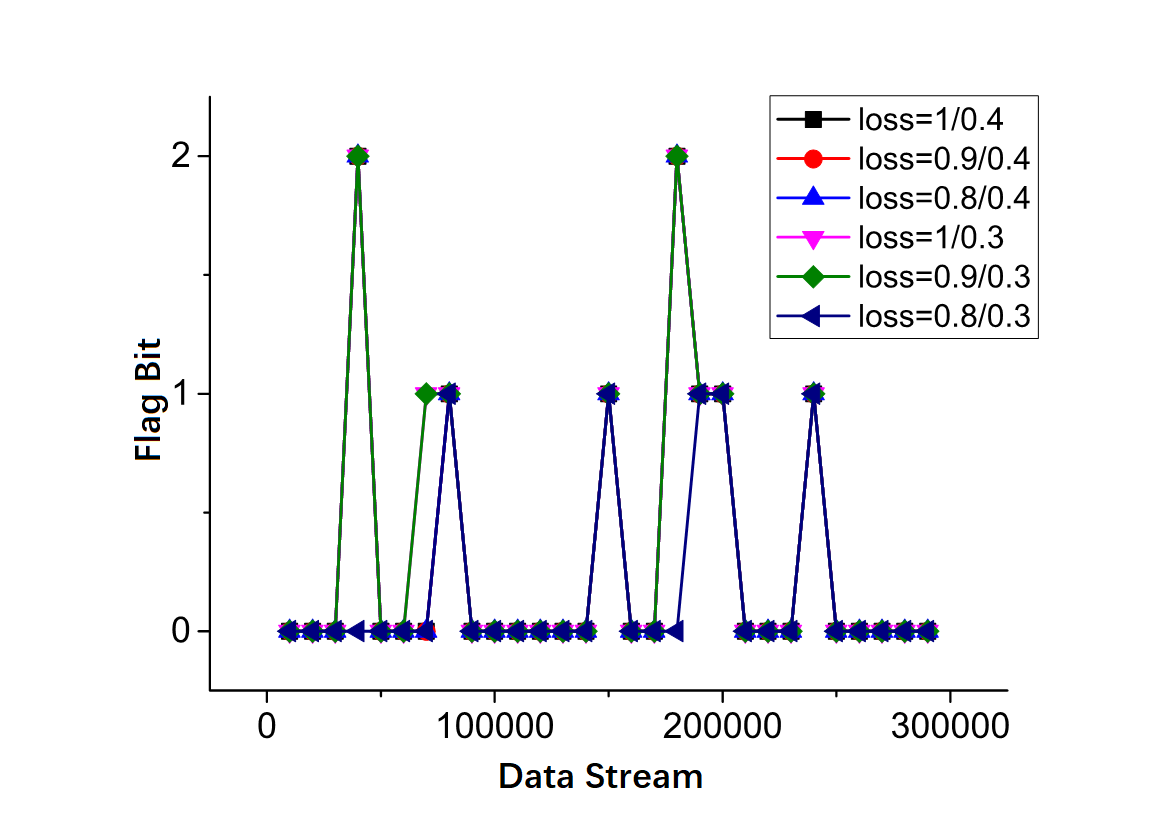}  
	\caption{The Impact of the Change Rate for Algorithm 2  }  
	\label{fig:9}   
\end{figure}
The similarity and the loss rate are tuned in a similar process for Algorithm 2. The experimental results are shown in Figures ~\ref{fig:8} and ~\ref{fig:9}. We have observed that when the similarity is 0.5 and the loss rate is 0.9/0.3, the updating is more effective. When the thresholds are moderate, the model can execute the three actions, update, discard or retain the model in a proper and balanced frequency.

From the experimental results described above, after tuning the prediction algorithm based on the data renewal model, the threshold is fixed at the similarity of 0.5, and the loss rate is 0.9/0.3. The experimental results show that the algorithm does not need to be tuned frequently and the fixed parameters can also achieve great experimental results.

\subsection{Results for the Time-series Yield Prediction Algorithm}
We test the effectiveness of the proposed model renewal strategy on the time-series yield prediction algorithm. It's an LSTM algorithm based on multi-variable tuning. The algorithm improves the traditional LSTM algorithm and converts the time-series data into supervised learning sequences utilizing their periodicity, so as to improve the prediction accuracy.
\begin{figure}[htb]
	\centering  
	\includegraphics[width=1.0\linewidth ]{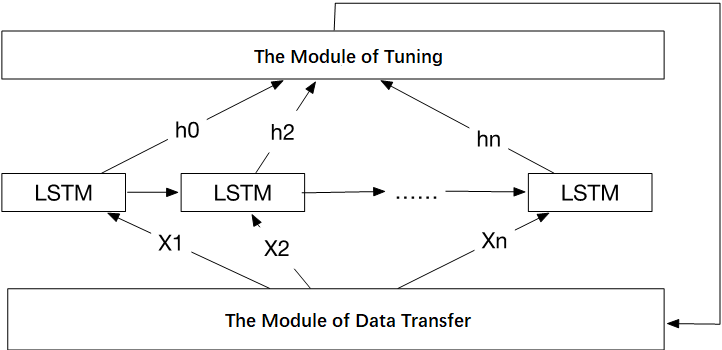}  
	\caption{ The Procedure of the Time-series Yield Prediction Algorithm }  
	\label{fig:10}   
\end{figure}
The LSTM algorithm based on multi-variable tuning is divided into three modules, a data transform module, an LSTM modeling module, and a tuning module. The data transform module converts the time-series data into a supervised learning sequence, and simultaneously searches for the variable sets which are most relevant and have the highest Y regression coefficient; the LSTM modeling module connects multiple LSTM perception to form an LSTM network; the tuning module adjusts the parameters according to the RMSE in each round, and returns the adjusted parameters to the data transform module for training iteratively. Through continuous iteration, the approximate optimization solution of the algorithm is obtained. The algorithm process is shown in Figure ~\ref{fig:10}. The following is the result of applying the data renewal model to the yield prediction algorithm based on time-series.


After determining the optimal thresholds, we selected different update frequencies, 10,000, 50,000 and 100,000 pieces of data for each batch, to verify the effectiveness of the model.
\begin{figure}[htb]
	\centering  
	\includegraphics[width=0.8\linewidth]{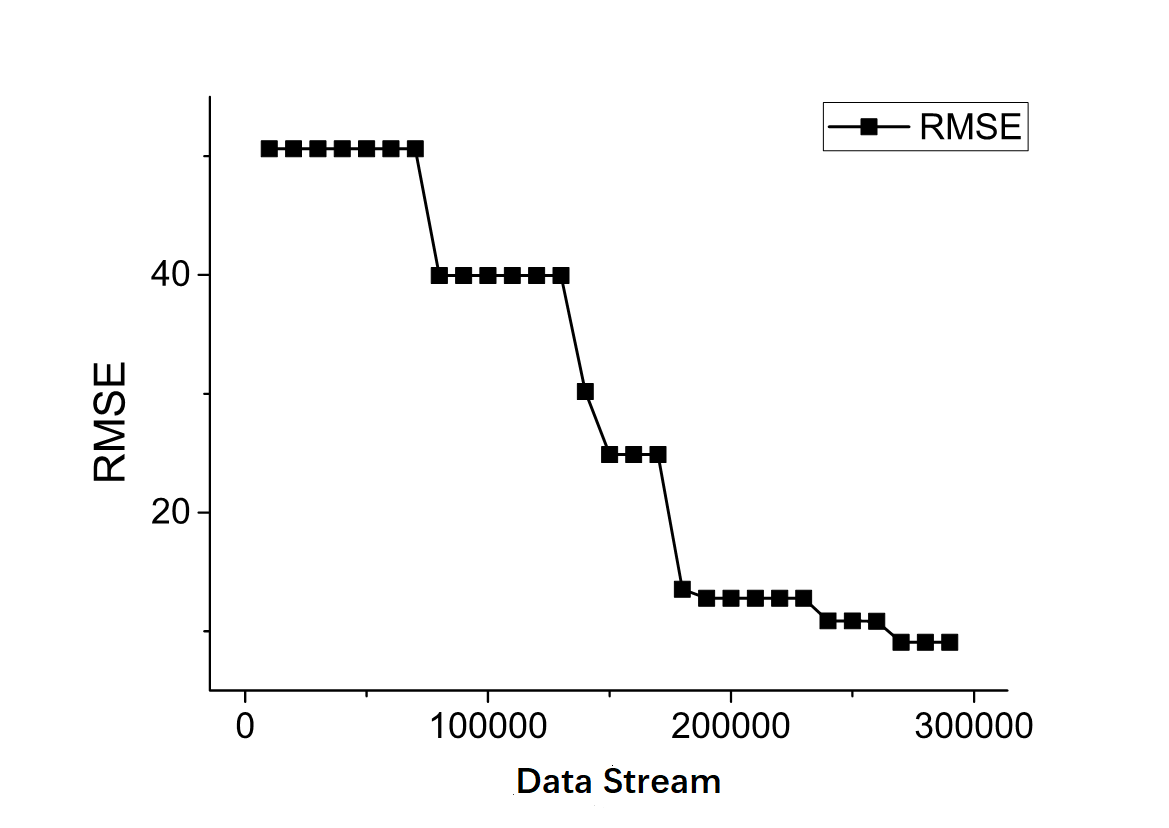}  
	\caption{The Result at a Frequency of 10000 pieces/batch}  
	\label{fig:11}   
\end{figure}
First, we set the data stream updating frequency to 10,000 pieces/batch. The experimental results are shown in Figure ~\ref{fig:11}. The RMSE changes with the continuously updating of the data stream. During this process, the model's RMSE remains unchanged (the flag is 0) or decreases (the flag is 1 or 2). As time goes by and the data accumulate, the RMSE of the model constantly decreases with an irregular frequency.

\begin{figure}[h]
	\centering  
	\includegraphics[width=0.8\linewidth]{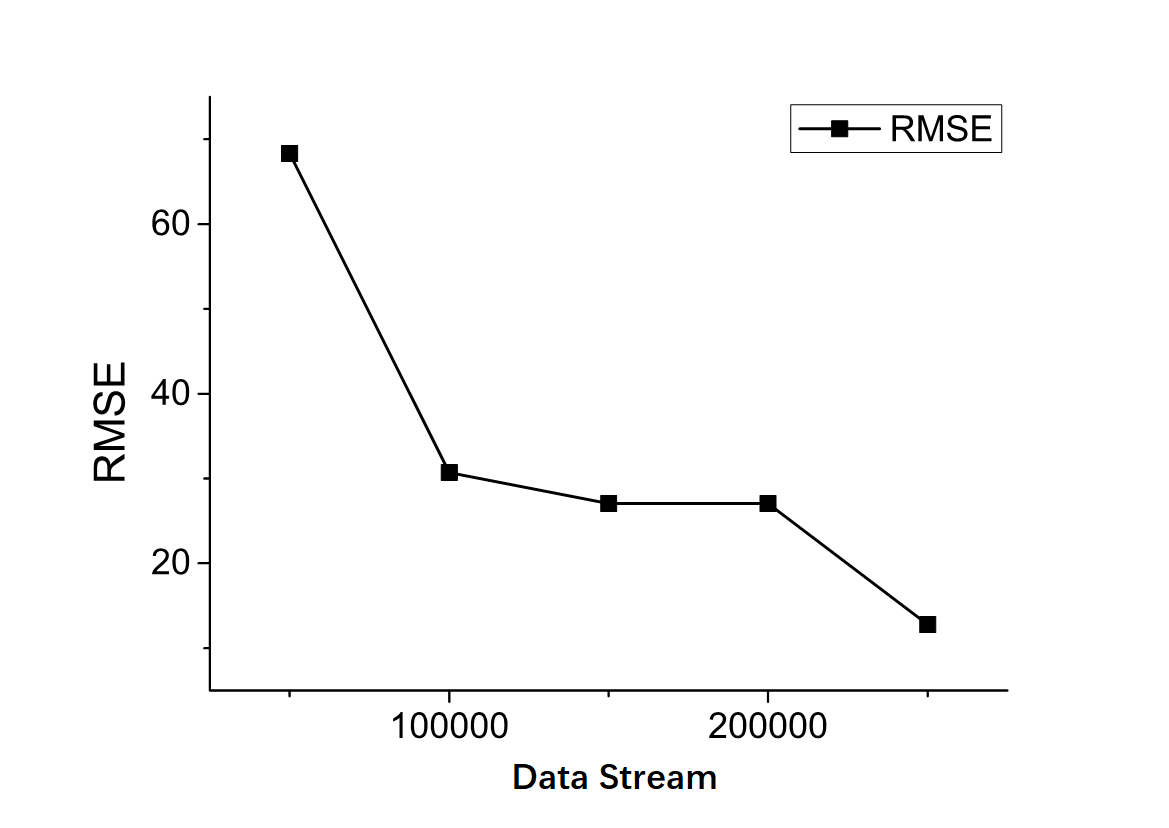}  
	\caption{The Result at a Frequency of 50000 pieces/batch }  
	\label{fig:12}   
\end{figure}
When the data updating frequency is 50,000 pieces/batch, the experimental results are shown in Figure ~\ref{fig:12}. It is found that when the updating frequency of the data is reduced, the RMSE suffers different degrees of reduction, which indicates that different data update frequencies will affect prediction accuracy. Nevertheless, when the amount of data tend to be consistent, overall, the RMSE can be reduced to a specific constant with less error.

We set the frequency for data updating to 100,000 pieces/batch. Let the original data be 100,000 pieces, and take the reduction degree of RMSE as its updating accuracy. As shown in Table 2, the algorithm can perform effective model updating in various frequencies. Moreover, the updating accuracy is up to 63.94\%.

\begin{table}[htb]
	\caption{Experimental Parameters of the Yield Prediction Algorithm }
	\label{Tab:bookRWCal}
	\centering
	\scriptsize
	\begin{tabular}{lp{1.2cm}p{1.1cm}p{1.6cm}p{0.7cm}p{0.8cm}p{1.5cm}}
		\hline
		No. & Data Volume & Similarity & Loss Rate & RMSE & Accuracy \\
		\hline
		1 & 100000& 0.34 &0.592957366 &30.17& -\\
		2&100000& 0.44 &0.795650783 &10.88& 63.94\% \\
		3&100000& 0.89 &0.122410221 &10.88& - \\
		\hline
	\end{tabular}
\end{table}

\subsection{Results for Transfer-learning-based Fault Prediction Algorithm}
In the problems of malfunction diagnosis and predictions, though there are various kinds of faults, the similarities of them can be utilized to adopt the transfer learning of malfunctions, so as to predict the errors efficiently and effectively. 
\begin{figure}[h]
	\centering  
	\includegraphics[width=1.0\linewidth]{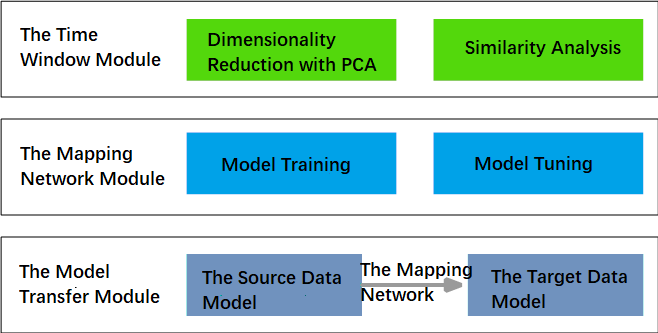}  
	\caption{The Schematic Diagram of the Transfer Learning Based Fault Prediction Algorithm  }  
	\label{fig:15}   
\end{figure}
The transfer-learning-based fault prediction algorithm mainly consists of three modules, the time window module, the mapping network module, and the model transfer module. The time window module preprocesses the data and conducts similarity analysis based on the time window. The mapping network module mainly uses the step of the similar region of time-series data to transfer the data, and construct the mapping network with the converted data. The model transfer module mainly utilizes the mapping network to transfer the trained model, which is prepared with the neural network method to construct a deep learning network. Figure ~\ref{fig:15} shows the structure of the algorithm.

In the experiment, we suppose that there are two devices, device $A$ with labeled data and device $B$ with unlabeled data. The output is the fault detection model $M$ of device $B$.
The update of the algorithm is mainly composed of two parts. For device $A$, the update target is the fault prediction model. For device $B$, if the data change, it is the mapping network module that should be updated. Therefore, our experiments are also divided into two parts.

For the prediction of device $A$, the original model is evaluated using the AUC (Area Under the Curve)
\cite{fawcett2006introduction}. When the update frequency of the data is 10,000 or 50,000 pieces/batch, the prediction accuracy AUC is about 0.97. When the frequency is 100,000 pieces /batch, the prediction accuracy AUC is about 0.958, and the loss is less than 1\%, so the model need not be updated after the initial modeling.

For device $B$ which gets new data, the updating process is more complicated. The mapping network module is updated first. Then the prediction model of device $A$ through the mapping network is trained. Finally, the prediction model of device $B$ is analyzed. The experimental results are similar to that of the time-series yield prediction algorithm. We set the update frequency to 10,000, 50,000 and 100,000 pieces/batch for the verification.
\begin{figure}[h]
	\centering  
	\includegraphics[width=0.8\linewidth]{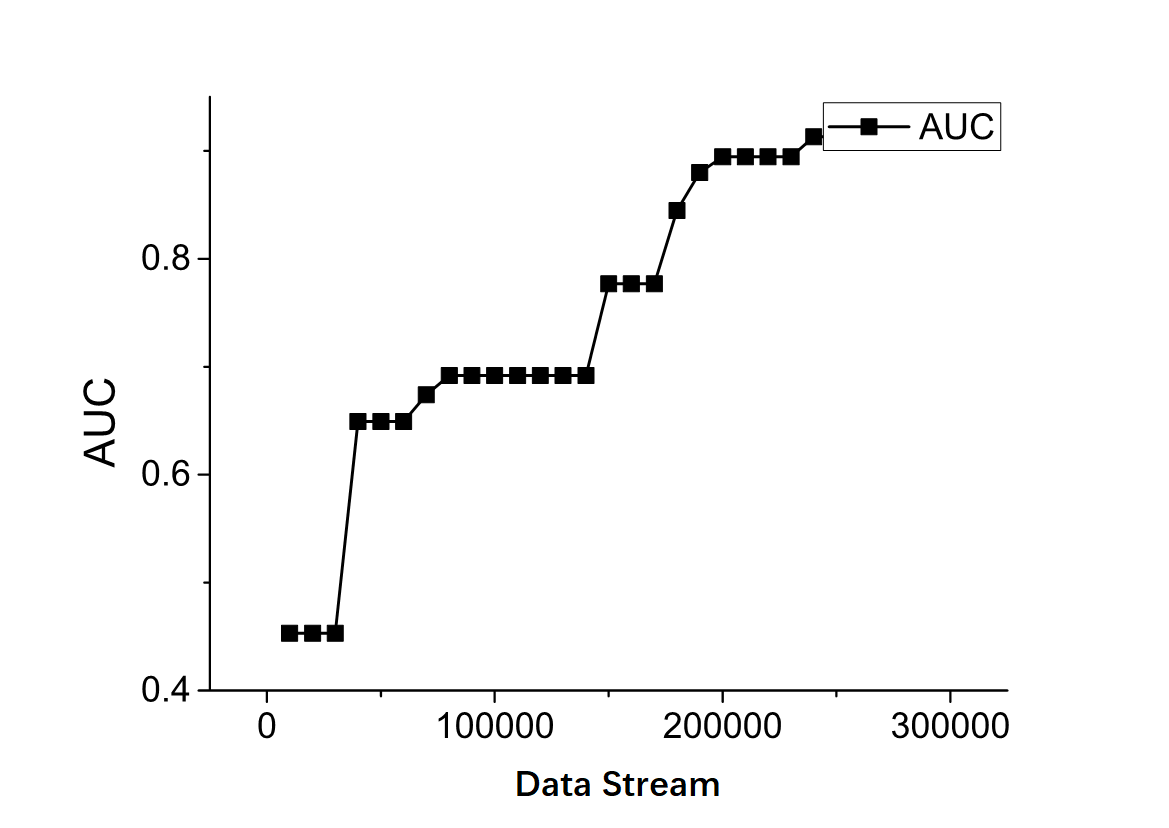}  
	\caption{The Result at a Frequency of 10000 pieces/batch }  
	\label{fig:13}   
\end{figure}
During updating, the model's AUC remains unchanged (the flag is 0) or increases (the flag is 1 or 2). The experimental results with a frequency of 10,000 pieces/batch are shown in Figure ~\ref{fig:13}. With the accumulation of data over time, the AUC of the model is continually increasing, which indicates better performance.

\begin{figure}[h]
	\centering  
	\includegraphics[width=0.8\linewidth]{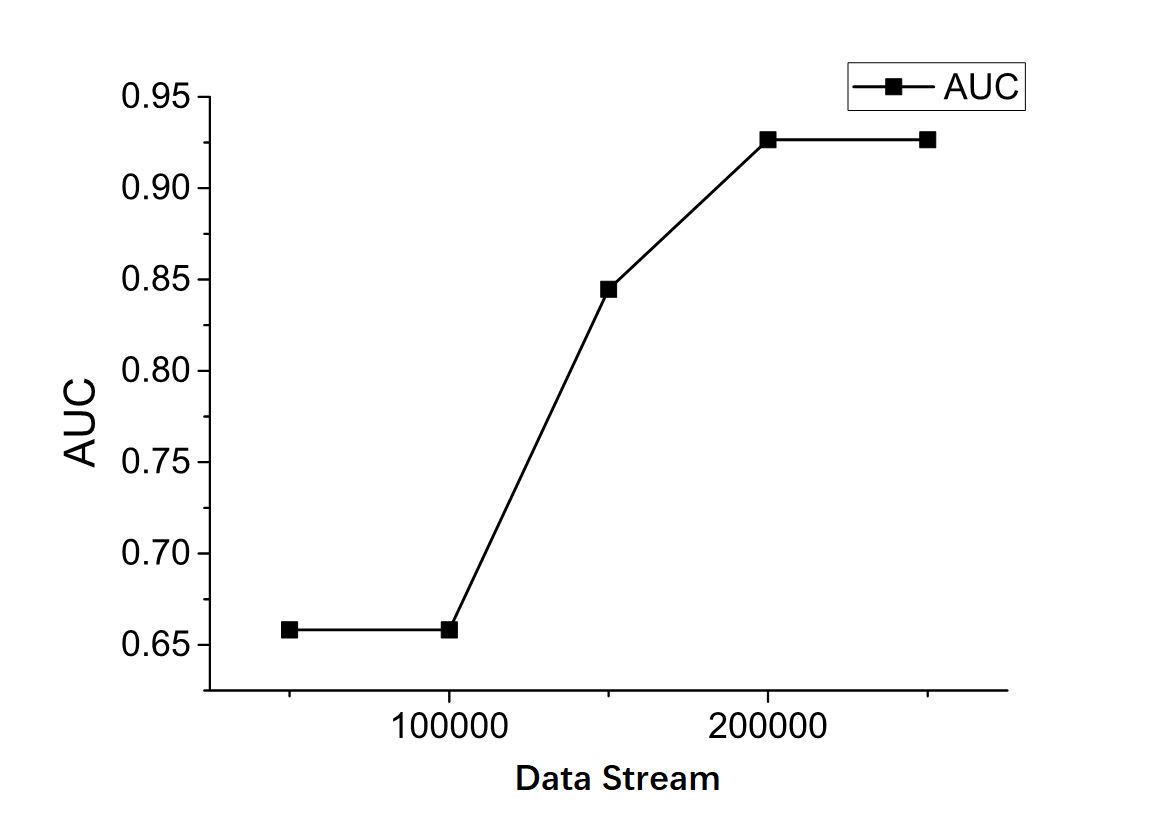}  
	\caption{The Result at a Frequency of 50000 pieces/batch  }  
	\label{fig:14}   
\end{figure}
When the data update frequency is 50,000 pieces/batch, the experimental results are shown in Figure ~\ref{fig:14}. It is observed that the degree of updating varies with different updating frequencies. When the amount of data tends to be consistent, the values of AUC will get stable.

We set the frequency for data update to 100,000  pieces/batch, and take the AUC as its accuracy. We observe that when the data similarity is low, there will be a more considerable degree of updating, as the results shown in Table 3. It demonstrates that the data renewal model can effectively update the existing model with the data stream in different updating frequencies.

\begin{table}[htb]
	\caption{Experimental Results for the Transfer-learning-based Fault Prediction Algorithm }
	\label{Tab:bookRWCal}
	\centering
	\scriptsize
	\begin{tabular}{lp{1.2cm}p{1.1cm}p{1.6cm}p{0.7cm}p{0.8cm}p{1.5cm}}
		\hline
		No. & Data Volume & Similarity & Loss Rate & AUC & Accuracy \\
		\hline
		4 & 100000& 0.33 &0.874255365 &0.68& -\\
		5&100000& 0.64 &0.571217426 &0.68& - \\
		6&100000& 0.23 &0.427696411 &0.91& 33.82\% \\
		\hline
	\end{tabular}
\end{table}

\section{Conclusions and future work}
In this paper, we propose a general data renewal model to assess the industrial data stream and set the thresholds to update the prediction model adaptively. It can by applied to some prediction algorithms to improve their performance. The effectiveness of the model and its significance to improving the accuracy of prediction are demonstrated by experiments on real-world industrial datasets and prediction algorithms. However, since it's a general model, the tuned parameters couldn't be adaptive in any different kind of problems. So self-adaptive tuning may be taken into consideration in the future work. And we only apply the model to two industrial prediction algorithms in our work, the model can be examined by more algorithms.
\bibliographystyle{IEEEtran}
\bibliography{sample-sigconf}
\end{document}